\documentclass[fleqn,10pt]{wlscirep}
    \usepackage[utf8]{inputenc}
    \usepackage[T1]{fontenc}
    \usepackage{lineno}
    \usepackage{array}
    \usepackage{float}
    \usepackage{nameref}
    \usepackage{xcolor}
    \usepackage{amsmath,amssymb}
    \usepackage{algpseudocode}
    \usepackage[ruled,vlined]{algorithm2e}
    \usepackage[acronym,toc]{glossaries}
    \usepackage{multirow}
    \usepackage{subcaption}
    \usepackage{longtable}

    \title{DiTEC-WDN: A Large-Scale Dataset of Hydraulic Scenarios across Multiple Water Distribution Networks}
    
    \author[1,*,$\dag$]{Huy Truong}
    \author[1,*,$\dag$]{Andr\'{e}s Tello}
    \author[1]{Alexander Lazovik}
    \author[2]{Victoria Degeler}
    \affil[1]{Bernoulli Institute, University of Groningen, Groningen, The Netherlands}
    \affil[2]{Informatics Institute, University of Amsterdam, Amsterdam, The Netherlands}
    
    \affil[*]{corresponding author(s): Huy Truong, Andr\'{e}s Tello (h.c.truong@rug.nl, andres.tello@rug.nl)}
    
    \affil[$\dag$]{these authors contributed equally to this work}

    \makeglossaries
    \setacronymstyle{long-short}

\newacronym{wdn}{WDN}{Water Distribution Network} 
\newacronym{adg}{ADG}{Automatic Demand Generator} 

\newacronym{hspo}{HSPO}{Hydraulic Sampling Parameters Optimization}
\newacronym{pso}{PSO}{Particle Swarm Optimization}

\newacronym{ubiqr}{UBIQR}{Upper Bound of the Inter-Quartile Range}

\newacronym{csv}{CSV}{comma-separated value}

\newacronym{wds}{WDS}{Water Distribution System} 
    \begin{abstract}
    Privacy restrictions hinder the sharing of real-world Water Distribution Network (WDN) models, limiting the application of emerging data-driven machine learning, which typically requires extensive observations. To address this challenge, we propose the dataset DiTEC-WDN that comprises 36,000 unique scenarios simulated over either short-term (24 hours) or long-term (1 year) periods. We constructed this dataset using an automated pipeline that optimizes crucial parameters (e.g., pressure, flow rate, and demand patterns), facilitates large-scale simulations, and records discrete, synthetic but hydraulically realistic states under standard conditions via rule validation and post-hoc analysis. With a total of 228 million generated graph-based states, DiTEC-WDN can support a variety of machine-learning tasks, including graph-level, node-level, and link-level regression, as well as time-series forecasting. This contribution, released under a public license, encourages open scientific research in the critical water sector, eliminates the risk of exposing sensitive data, and fulfills the need for a large-scale water distribution network benchmark for study comparisons and scenario analysis.
    \end{abstract}
    
\begin{document}
    
    \flushbottom
    \maketitle
    
    \thispagestyle{empty}
    
    
    \section*{Background \& Summary}  
    
    

    	\glspl{wdn} are considered critical infrastructures as they provide clean and safe water to humans, which is one of the Sustainable Development Goals proposed by the United Nations. Water providers have to deal with critical challenges during the design, planning, and management phases of a \gls{wdn} in order to fulfill this goal, such as adaptability and robustness to an ever changing environment. Climate, consumer behavior, aging infrastructure, failures, all can lead to drastic changes in the conditions under which \glspl{wdn} must continue working adequately.   Monitoring of \gls{wdn} operations plays an important role in guaranteeing the water supply. The state of the network must be known at any given time to prevent unwanted situations, e.g., pipe leaks. 
    	
    	Hydraulic modeling has been the most straightforward approach for practitioners to simulate the \gls{wdn} dynamics and aid design, planning and management. While pure physics-based hydraulic modeling is still being commonly used in the water domain, water engineering research and practice are experiencing a shift towards hybrid data-driven approaches. Such approaches combine the power of physics and mathematical simulation tools with data-driven deep learning models to solve water engineering problems. Paradoxically, while data are the key to such approaches, \gls{wdn} operation data are scarce and seldom shared among practitioners and researchers due to privacy, safety and other domain related constraints \cite{brumbelow2007virtual, sitzenfrei2013automatic}. A notable example are nodal demand patterns. Demand is one of the most important inputs for solving the \gls{wdn} hydraulics \cite{giustolisi2012demand}. Surprisingly, it is one of the inputs that is rarely found in the \gls{wdn} asset description files. It is common to find just a few demand patterns reused many times on several nodes in the network \cite{ostfeld2012battle}, or not demand patterns at all \cite{reca2006genetic}. The stochastic nature of water demand explains some of the uncertainties found in \glspl{wdn} \cite{zanfei2023shall}, which should be properly modeled and considered in the simulations \cite{cassiolato2024water}. Hence, reusing the same demand patterns on multiple nodes assumes that several users/consumers have exactly the same water consumption behavior, which is unrealistic. This not only harms the robustness of the models to uncertainties, but also limits the variety of the data, which is especially important for deep learning data-driven approaches.
    		
    	Benchmark datasets for \gls{wdn} data analysis are, in fact, very limited \cite{vrachimis2018leakdb, tello2024gida}. LeakDB \cite{vrachimis2018leakdb} is a dataset commonly used in research at the moment, but it only includes a single small \gls{wdn} with limited variability of scenarios, as explained in Section Technical Validation. This limits the diversity of the data for training data-driven models. More commonly found in practice is a collection of static asset descriptions of water networks, and different algorithms and implementations for data generation from them. Researchers and practitioners working on data-driven approaches for water engineering lack data to train their models, and count on static asset descriptions of the \gls{wdn} rather than operational data. Those asset descriptions are represented as configuration files, which serve as input to physics-based mathematical tools to simulate the data required for data-driven models' training. Although the simulation of \gls{wdn} hydraulics from well-defined configuration files seems to be straightforward, it is a cumbersome process that involves expert knowledge, time-consuming models' calibration, uncertainties, computational complexity, among other challenges. Moreover, such configuration files only allow practitioners to simulate the \gls{wdn} states determined by the input parameters explicitly specified. Hence, if new data for a different \gls{wdn} is needed, or a different condition in the input parameters needs to be evaluated, the whole process has to be repeated from scratch. 
    	
    	The aim of this work is to support the shift towards data-driven approaches for \gls{wdn} data analysis. We provide a multi-purpose dataset generated based on 36 publicly available \glspl{wdn}, which includes 228 million of network state snapshots,  when operating under normal conditions. The synthetic nature of the data eliminates the privacy and safety concerns, facilitating data sharing among researchers, or even within the commercial sector, without any risks. The examples of tasks supported by these data are surrogate modeling, state estimation, and demand forecasting. The data provided include all the inputs used for the simulations and their respective outputs, which allows researchers to work on surrogate modeling of physics-based mathematical simulation tools. The large number of provided snapshots allow practitioners to work on state estimation models. The data include unique demand patterns per node, facilitating demand forecasting.

    \section*{Methods}
    
    
    
    
    
    
     
    
    \subsection*{Data acquisition}
    	\label{sec:data_acquisition}
    	The described synthetic dataset was created based on publicly available \glspl{wdn}. In order to achieve this, we collected data related to the topology and the physical properties of the networks' components. As mentioned before, such information is available as configuration files describing the assets of the \glspl{wdn}. Initially we collected the asset description data of 55 \glspl{wdn}. In those initial files, we found duplicated data related to the same \gls{wdn} but under different names. We also found configuration files with unreadable characters which did not allow a proper data reading. After a data depuration process we included 36 \glspl{wdn} in our final dataset. The full list of the \glspl{wdn} included in our dataset, and their main components, is shown in Table~\ref{tab:baselines}.
    	
    	Our data is generated using the EPANET \cite{rossman1999epanet} and WNTR \cite{klise2018overview} physics-based simulation tools, which allow to run simulations of the hydraulic behavior of \glspl{wdn}. These tools are widely used by researchers and practitioners in the water domain. All collected configuration files are represented in EPANET input file format (.inp). 
    	
    	The input file contains the metadata about the \gls{wdn} and the description of the components of the network, the system's operation, water quality, and other options used at simulation time. The file is organized in sections, where each section begins with a keyword enclosed in brackets. For example, the sections related to the network components include: [TITLE], [JUNCTIONS], [RESERVOIRS], [TANKS], [PIPES], [PUMPS], [VALVES], and [EMITTERS]. The sections related to the system's operation include: [CURVES], [PATTERNS], [ENERGY], [STATUS], [CONTROLS], [RULES], and [DEMANDS]. The complete description of the input file format can be found in the EPANET 2.2 User Manual \cite{rossman2000epanet}.
    	
    	In the context of this work, the EPANET input file represents the input to the data generation process. Accordingly, each section represents a collection of parameters that needs to be optimized in order to obtain a simulation outcome that is considered to be a valid state of the network. The [PATTERNS] section is used to specify the water consumption patterns associated with each junction node. The pattern is represented as a list of values, where each element of the list represents the water consumption at time step $t$. Another important section is [TIMES], there we can specify the duration of the simulation and the time step, i.e., the sampling rate of the simulation's outputs. 
    	
    	At runtime, the simulation generates a set of outputs corresponding to time step $t$, which is the state of the network at such given time. In our work, each network state is called a \textit{snapshot}. The collection of snapshots that span the entire duration of the simulation is called a \textit{scenario}. This dataset includes 10 \glspl{wdn} where each scenario spans 24 hours, and 26 \glspl{wdn} where each scenario spans 1 year, with a 1 hour time step in both cases. The complete dataset comprises 1,000 scenarios per network, which represent 228-million snapshots of water networks' states.

    

    \subsection*{Data generation}
    Following the network collection, we present a data generation scheme to synthesize simulated data in individual water networks. Overall, the scheme involves three subsequent steps: Data Preprocessing, Hydraulic Parameter Optimization, and Simulation. The first step filters targeted simulation parameters and collects statistics across available networks. Both are then fed into an optimization algorithm to determine the sampling strategy and corresponding bounds for specific parameters. The last step plays a role in sampling concrete values, performing simulation, and encapsulating the data in a compressed format. We now describe each step in detail.
    
    \paragraph{Preprocessing step}
            
    A static network description from an input file, described in Section Data Acquisition, contains useful simulation-oriented data and irrelevant information, such as titles, labels, and water quality parameters. Since this study focuses on hydraulic-related parameters, it is crucial to filter and refine only this specific data before proceeding to the next stage. Table \ref{tab:parameters} indicates selected parameters and their corresponding information. For some parameters, the original measurement units vary depending on the geographical region of each water network. For example, demand is measured in liters per second (LPS) in \textit{hanoi} \gls{wdn}, but in gallons per minute (GPM) in \textit{ky8} \gls{wdn}. For the sake of consistency, they are converted to the corresponding International System of Units (SI system) using the wrapper tool WNTR~\cite{klise2018overview}.
    
    These selected parameters are then stored in a YAML configuration file. It is similar to the input file but contains essential metadata for both optimization and simulation phases, such as computed duration, time step, and names of skipped nodes. The configuration also records the sampling strategy and bounds for available parameters in a specific network. This metadata is included in the final delivery for reproducibility purposes.
    
    Besides the configuration files, another type of information is computed by the profiler, calculating statistics of those 38 parameters collected from original water networks. For each parameter, the profiler captures the \textit{minimum}, \textit{maximum}, \textit{mean}, \textit{standard deviation}, \textit{first quartile}, \textit{third quartile}, \textit{parameter dimension}, and \textit{the number of components} that can obtain this parameter. The statistics are computed for each baseline network and, additionally, for the global network representing the overall perspective. We leveraged them to 1) determine the sampling range and size and 2) perform data imputation in case of missing values in the following step. 
    
    \paragraph{\acrfull{hspo}} 
    
    Consider a \gls{wdn} with $n$ nodes and $m$ links, where each node and edge can obtain three types of parameters: static, pattern, and curve. The static parameter is a scalar or categorical value assigned per component, such as \textit{elevation} or \textit{status}. A pattern is a time series that typically changes throughout the scenario (e.g., \textit{junction input demand}, \textit{head pattern}). A curve defines the relationship between two measurements, such as a \textit{pump curve}, which reflects a pump's operating capacity based on flow rate and head. Assume a node has $s_n$ static parameters, $p_n$ patterns, and $c_n$ curves, each with a maximum length $l_n$, with corresponding parameters for edges represented by $s_e, p_e, c_e, l_e$. Given a simulation duration $d$, a simulation candidate lies in a space of $s_n + p_n d + c_n l_n + s_e + p_e d + c_e l_e$ 
    dimensions. Given this high dimensionality, we consider an alternative approach: identifying a sampling strategy to define appropriate values per parameter to generate a simulation candidate while reducing the search space but preserving the data diversity. We call this the \gls{hspo} problem.
    
    In particular, \gls{hspo} aims to identify stable pairs of sampling strategies and value bounds for each hydraulic parameter of all components. In other words, given a baseline \gls{wdn}, the goal is to model numerous network variants and validate their parameters to ensure data stability within a specified time frame. There are two time frames, corresponding to the two dataset types: short-term and long-term. The short-term dataset includes scenarios observed over a 24-hour period, while the long-term dataset covers scenarios with a span of 1 year. Both use an hourly time step for sampling. 
    Before diving into details, we outline the following potential sampling strategies to determine the value range of a specific parameter:
    \begin{itemize}
        \item \textbf{Keep}. Following the principle “Doing nothing is better than doing anything”, this strategy preserves the parameter's state as in the baseline network. This approach significantly reduces the search space and, therefore, mitigates the optimization complexity~\cite{donninger1993null}
        
        \item \textbf{Series}. This strategy applies an existing series of a particular parameter across all components. For instance, \textit{pump curve pattern} can be retrieved in the pump manual supplied by the manufacturer and applied to every pump curve within the networks. The value is then shared across all scenarios.
    
        \item \textbf{Sampling}. Given a predefined range $[min, max]$ of a particular parameter, we uniformly sample a new value for a hydraulic parameter per component. This approach ensures that every component has its own distinct value. For patterns and curves, this strategy leverages statistics from the profiler to sample series accordingly.
        
        \item \textbf{Perturbation}. For a parameter, we gather the mean and standard deviation from the baseline \gls{wdn} and sample from a Gaussian distribution. This strategy is beneficial when the parameter's value is unavailable in the target network, allowing us to use values from the global perspective.
    
        \item \textbf{Factor}. We sample a scale and bias to apply a linear transformation to existing values gathered from the baseline network. This approach ensures consistency, which is essential for certain parameters. For example, three adjacent pipes should have similar diameters. In such case, the \textbf{Factor} serves as a potential strategy, while \textbf{Sampling} and \textbf{Perturbation} cause pipe bottleneck as a modeling anomaly in practice. 
    
        \item \textbf{Substitute}. It randomly selects an existing value of the target parameter and shares it with all components. This approach also injects minor noise into the values to maintain diversity. Similar to the Factor method, it respects consistency in modeling.
    
        \item \textbf{Terrain}. This is a special strategy applicable to junctions' elevation. In particular, we employ the Diamond Square algorithm with proper noise to generate a 2D height map~\cite{fournier1982diamondsquare}. Given the nodal coordinates from the input file, we project the network onto the map to obtain new elevations. 
    
        \item \textbf{\gls{adg}}. This sampling strategy is specially tailored for junctions' input demands, the most crucial but scarce parameter. Due to its importance, Section Automatic Demand Generator is dedicated to describing this approach. 
    \end{itemize}
    
    For each target \gls{wdn}, a default blueprint configuration is set up as follows: \textbf{\gls{adg}} for junction input demand, \textbf{terrain} for junction elevation, \textbf{factor (substitute)} for pipe diameter, and \textbf{keep} for all remaining parameters. Following this, the configuration is fed into a \gls{hspo} algorithm to iteratively refine the sampling strategy and sampling values for all parameters until convergence. 
    \paragraph{\acrfull{pso}.}
    
    Assume that \textbf{sampling} is the default generation strategy, each parameter needs a lower bound $lb$ and upper bound $ub$ to construct the sample space. This yields a total of $2D$ sampling parameters in the \gls{hspo} problem. Here, this problem is solved by \gls{pso}~\cite{kennedy1995pso}. 
    
    Mathematically, \gls{pso} opts to construct a solution $\mathbf{X} \in \mathbb{R}^{D \times 2}$, representing a sampling configuration. This configuration drives a function $gensim:  \mathbb{R}^{D \times 2} \to \mathbb{R}^{out \times d}$ to yield a scenario corresponding to $out$ measurements, each as a $d$-length time series.
    Internally, $gensim$ implicitly solves a system of equations~\cite{klise2018overview} and typically produces a large batch of diverse scenarios in practice. From this perspective, only $N_{\text{cases}}$ created scenarios are considered to evaluate a sampling solution. Nevertheless, their measurements could exhibit anomalies, such as negative pressure or time inconsistency. As the dataset is expected to be clean, this violates our assumption. To alleviate this, we form a set of rules $R = \{r_1, r_2...\}$ in which each rule judges whether simulated outcomes are valid. 
    
\begin{figure}[ht]
        \centering
        \includegraphics[width=\linewidth]{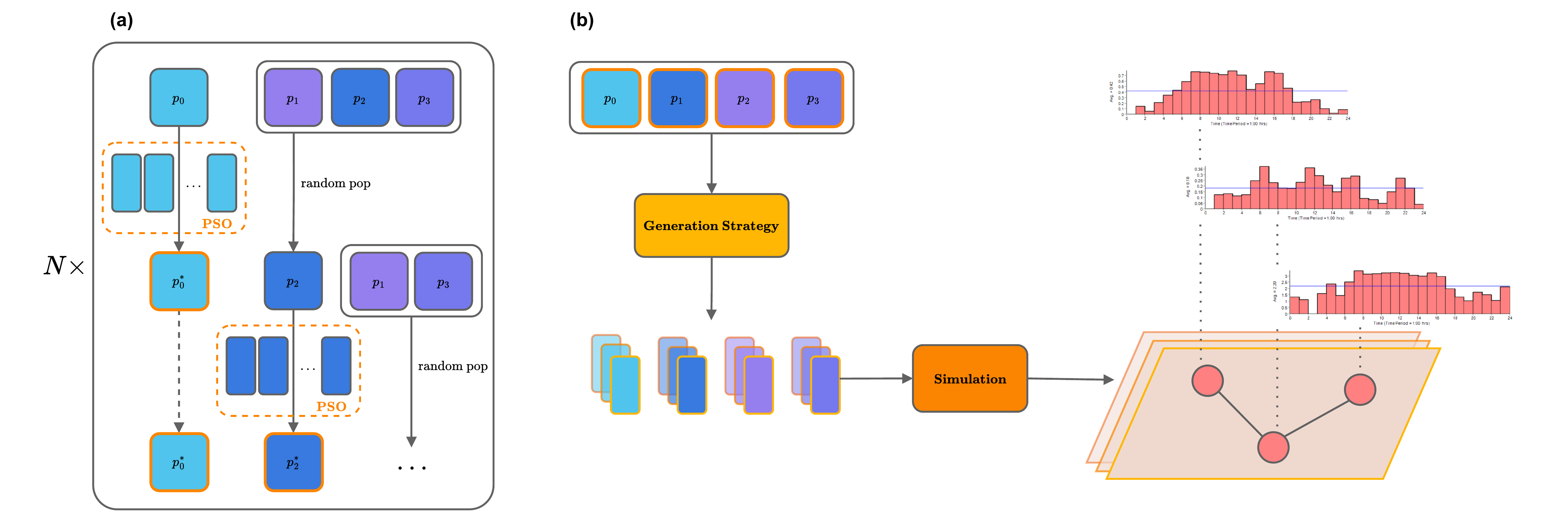}
        \caption{\textbf{Illustration of the dataset generation.} The left figure (a) shows a divide-and-conquer PSO optimizing a strategy's configuration. The right figure (b) depicts the usage of the optimized configuration to sample parameter sets and simulate diverse scenarios with unique characteristics (e.g., per-node demand patterns).}
        \label{fig:pso_and_simgen}
    \end{figure}

    Formally, a binary function $validate: \mathbb{R}^{D \times 2}  \to \{1,0\}$ is defined as follows:
    \begin{equation}
    \text{validate}(X) =  \left\{
        \begin{array}{ll}
            1 & \text{if } \forall r \in R, r(\text{gensim}(X)) = true \\
            0 & \text{otherwise}
        \end{array}
        \right.
    \end{equation}
    Nevertheless, empirical trials indicate that simultaneously optimizing all sampling parameters struggles to converge, more frequently for long-term cases ($t=8,760)$. This can be attributed to the large search space. To mitigate this, we facilitate \gls{pso} with a divide-and-conquer approach. As shown in Figure \ref{fig:pso_and_simgen}, \gls{pso} considers a sampling set of a particular parameter while maintaining the fixed state of other sets at every timestep. This isolation reduces the complexity and makes \gls{pso} more manageable than addressing all parameters simultaneously. After an iteration, the updated optimal value for the selected parameter is locked in the remainder of the epoch. A new \gls{pso} is then executed to optimize a random candidate from the parameter list, iterating until the list is empty. This process repeats across multiple epochs until the maximum number of epochs is reached or the intermediate solution is desired.
    
    At each iteration, a sampling solution could be formed as a concatenation of the latest optimized and other sets. We evaluated the ``goodness'' of this solution by defining a fitness function $f_{success}: 
    \mathbb{R}^{D\times2} \to \mathbb{R}$ computing the average success rate over $N_{\text{cases}}$ generation cases:
    \begin{equation}
    f_{success}(X) = \frac{ \sum^{N_{\text{cases}}}_{i=1}{\text{validate}(X)} }{ N_{\text{cases}}}
    \end{equation}
    Considering the stochastic nature, we set $N_{\text{cases}}$ to $100$ to estimate the goodness of each sampling solution. However, merely relying on $f_{success}$ leads to a collapse of the solution since particles tend to shrink in a local optimum, which is unrealistic and results in poor generalization. For instance, in one case of \textit{junction elevation}, \gls{pso} proposed a narrow sampling range of $[0.12, 0.12]$, resulting in flat terrain. To restrict such cases, a customized fitness function was designed.
    
    Assume a particle $i$ has its position represented as a solution $x_i \in \mathbb{R}^{D\times 2}$, we designed a fitness function $f_{pso}: \mathbb{R}^{D\times 2} \to \mathbb{R}$ as follows:
    \begin{equation}
        f_{pso}(x_i) =  f_{success}(x_i) \biggl( \alpha f_{ubiqr}(x_i)) + (1-\alpha)f_{range}(x_i) \biggr)
    \label{eqa:pso_loss}
    \end{equation} 
    where $\alpha$ is a hyper-parameter balancing the two auxiliary criteria: diversity indicator $f_{ubiqr}$, and range expansion measurement $f_{range}$. While the success ratio $f_{success}$ still plays a crucial role in assessing goodness, we encourage \gls{pso} to find optimal solutions beyond the baseline scenario. The $f_{ubiqr}$ computes the \gls{ubiqr}, a statistical measure of the spread of populations~\cite{vinutha2018iqr}. In this study, we compare the \gls{ubiqr} of \textit{junction output demand} between a generated case and the baseline, denoted as $y_i$ and $y_{bl}$. For the sake of brevity, we implied a simulation executed before computing this fitness (i.e., $y_i = sim(x_i)$). Mathematically, the comparison can be written as:
    \begin{equation}
        f_{ubiqr}(y_i) = \frac{\text{UBIQR}(y_i)}{\text{UBIQR}(y_{bl})}
    \end{equation}
    The last fitness $f_{range}$ encourages the expansion of the sampling range. For \textbf{Sampling} strategy with two normalized value bounds ($v_{min}$, $v_{max}$), the calculation is expressed as:
    \begin{equation}
        f_{range} = |v_{max} - v_{min}| 
    \end{equation}
    Using the combination loss given in Equation \ref{eqa:pso_loss}, the modified \gls{pso} algorithm iteratively evaluates and ``exploits'' values of the sampling set of a specific hydraulic parameter while holding the latest states of other parameters constant over an extended timeframe. In addition, as shown in Figure \ref{fig:pso_and_simgen}, parameter permutation introduces uncertainty, allowing \gls{pso} to explore solutions within a dynamic landscape. This strategy enables us to retrieve optimal sampling sets of hydraulic parameters for all available networks. These sets are stored in corresponding networks' configurations and, therefore, leveraged by a simulation to produce data points on a large scale.
    
    \paragraph{Simulation}
    The subfigure (b) of Figure \ref{fig:pso_and_simgen} illustrates the simulation workflow. Overall, the entire workflow leverages multi-core processing powered by a high-performance computing cluster. In particular, following the optimization process, the optimal sampling set associated with its strategy was transferred to the Generator where we sampled actual simulation parameters. These parameters were batched and passed through a Simulator. The Simulator simulated outcomes and evaluated scenarios based on a predefined set of rules. Finally, the input and output parameters of the validated scenarios were encapsulated in a compressed file.
    
    \subsubsection*{Automatic Demand Generator} 
    \label{sec:adg}

    The \gls{adg} algorithm aims to generate the junctions' demand patterns for each node in a \gls{wdn}. The demand pattern is defined following an additive model of three components: a daily pattern, a yearly seasonal pattern, and noise, as expressed in Equation \ref{eq:demand_pattern}. 
    
    \begin{equation}
    	\label{eq:demand_pattern}
    	D = \text{daily}(x_t) + \text{yearly}(x_t) + \epsilon_t: ~ t \in T
    \end{equation}
    
    \noindent where $D$ is the demand pattern of each node in the network, $\text{daily}(x_t)$ is the daily pattern, $\text{yearly}(x_t)$ is the yearly seasonal pattern, and $\epsilon_t$ is white noise. The demand patterns generated are a multipliers time-series, i.e., a factor that is multiplied by the \textit{base demand} of each node specified in the configuration file of each \gls{wdn}. The generated time-series are normalized in the range $[0,1]$.
    
    \paragraph{Daily Pattern.} The daily pattern defines the water consumption per day based on consumption profiles: household, commercial, extreme-demand, zero-demand. The consumption profiles are determined by splitting the 24-hour of a day into four 6-hour segments. Thus, starting at midnight, these segments represent the water consumption from 00:00 to 06:00, 06:00 to 12:00, 12:00 to 18:00, and 18:00 to 00:00. Each segment is assigned either a low, medium or high consumption. The range for low, medium or high consumption is defined by lower and upper bounds determined at random. Thus, from N random numbers in the range $[0.00, 1.00]$ we compute the quantiles Q1, and Q3. Then, the low consumption goes from $[0.00, Q1)$, the medium consumption goes from $[Q1, Q3)$, and high consumption is in the range $[Q3, 1]$. For example, the household profile is represented as (low, high, medium, low). It is assumed a low consumption between midnight and six in the morning, with a peak consumption in the morning when people are preparing for work. Then, after noon, the demand gradually decreases during the day because people are at work, and finally the demand is low again at the end of the day when people are going to bed. In a similar way the commercial profile is defined as (high, high, high, medium). In this case, assuming a high consumption most of the time with a small decrease at the end of the day. 
    
    Using the consumption ranges described before, we generate random samples for each of the 6-hour segments. The number of $samples\_per\_hour$ is determined based on the sampling frequency ($time\_step$)  defined in the configuration file. Those 6-hour segments are then concatenated to compose the 24-hour corresponding to one day. Then, these 24-hour samples are repeated to span the entire \textit{duration} of the demand pattern. The daily demand pattern is generated using the periodic function described in Equation \ref{eq:dts}.
    
    \begin{equation}
    	\label{eq:dts}
    	\text{daily}(x_t) = cos(x_t) + sin(x_t) + z_t: ~ t \in T
    \end{equation}
    
    \noindent where the $cos(.)$ and $sin(.)$ terms introduce the daily periodicity in the time-series, $x_t$ represents the previously generated random sample at time $t$, and the $z_t$ term represents white noise. The noise component guarantees that each repetition of the 24-hour pattern along the time-series is not a fidelity copy of the previous one. Finally, we use the Savitzky-Golay filter \cite{savitzky1964smoothing, luo2005savitzky} to smooth the generated time-series.
    
    After the consumption profiles are defined, they have to be assigned to each node in the network. Hence, we need to determine which nodes belong to household profile and which ones to commercial. Domain knowledge indicates that commercial nodes are grouped in certain regions of the \gls{wdn}. In order to resemble this characteristic, we propose to cluster the nodes into two main groups household and commercial. The clusters are computed using the Louvain community detection algorithm, a heuristic approach that maximizes the modularity of the network \cite{blondel2008fast}. This algorithm works in two phases. In the first phase, each node $i$ is isolated and belongs to a community $C$. Then, the modularity gain is computed after each node is moved to its neighbor communities. If there is no positive gain in modularity, the node remains in its original community. This phase is repeated until no individual move can improve the modularity. For directed graphs, the modularity gain is computed as follows~\cite{blondel2008fast, traag2019louvain, dugue2015directed}:
    
    \begin{equation}
    	\Delta Q = \frac{k_{i,\text{in}}}{m} - \gamma \frac{k_{i}^{\text{out}} \cdot \Sigma_{\text{tot}}^{\text{in}} + k_{i}^{\text{in}} \cdot \Sigma_{\text{tot}}^{\text{out}}}{m^2}
    \end{equation}
    
    \noindent where $m$ is the size of the network, $\gamma$ is the resolution parameter which controls the size of the communities \cite{newman2016equivalence}, $k_{i}^{\text{out}}$, $k_{i}^{\text{in}}$ are the outer and inner weighted degrees of node $i$, and $\Sigma_{\text{tot}}^{\text{in}}$, $\Sigma_{\text{tot}}^{\text{out}}$ are the sum of in-going and out-going links incident to nodes in 
    community $C$. 
    
    In the second phase, the communities found in the previous step become nodes in the network, and the weights of links in the new graph are the sum of the weight of the links between nodes in the corresponding communities. Then the whole algorithm is applied again. The algorithm stops when no modularity gain is achieved or when the modularity is lower than certain \textit{threshold}.  
    
    At this stage, we have coherent communities within each \gls{wdn}. Now, we need to define the number of nodes from those communities that will be assigned to either commercial or household profiles. According to the statistics provided by the association of water companies in the Netherlands, about 28\% of the users belong to the commercial sector \cite{vewin2015dutch, vewin2017dutch, vewin2022dutch}. We randomly choose the $percentage\_commercial$ from the range $(0.25, 0.35)$. This allows to resemble commercial consumption profiles in other countries around the world. We set the number of nodes that will be assigned to the commercial consumption profile as $num\_nodes\_commercial = floor(percentage\_commercial \times total\_number\_of\_nodes)$. After that, we iterate the communities found in the previous stage and sequentially assign the nodes in each community to the commercial consumption profile until we reach the $num\_nodes\_commercial$. Finally, the remaining nodes will be assigned to household profile at this stage. While household and commercial profiles are self-explanatory, extreme and zero-demand are a special type of consumption profiles.
    
    The extreme-demand is a special case for some nodes with a very high water consumption. Thus, the extreme-demand profile is represented as (high, high, high, high). Usually an extreme node represents a group of nodes, commonly external to the water network, but also connected to it. We set the $extreme\_dem\_rate = 0.02$, i.e., 2\% of the scenarios will have nodes whose demand is always high. In addition, we limited the number of nodes per scenario that can have extreme demand values, specifically we set $max\_extreme\_dem\_junctions=2$. Domain knowledge can help to determine this parameter if the number of extreme nodes is known beforehand. The nodes to be assigned an extreme-demand profile are chosen at random, and excluded, from the nodes in the household or commercial profile. Then, for these nodes, the demand is randomly generated in the range $[Q3, 1]$, as described before. 
    
    The zero-demand is another special case that represents nodes that do not consume water, but which are part of the network. Thus, these nodes always have a zero-demand value. These nodes are used for monitoring and control of the network operation, or they are modeled due to a planned expansion of the network. We set the $zero\_dem\_rate = 0.05$, i.e., 5\% of the scenarios will have nodes whose demand is zero. Likewise, 5\% of the total number of nodes in the \gls{wdn} will be assigned the zero-demand profile. Alternatively, the $zero\_dem\_rate$ can be set to the ratio between the number of nodes in the baseline network whose \textit{base demand} is zero with respect to the total number of nodes; and accordingly, the number of nodes belonging to this profile. The zero-demand nodes are chosen at random, and excluded, from the remaining household or commercial profiles.
    
    The presence and use of both, extreme-nodes and zero-demand nodes, at modeling \glspl{wdn} are seen in the  baselines and also confirmed by experts in the water management domain. Including these two profiles in the generated data enables to cover a wider range of pressures and demands compared to the baselines. Otherwise, if the baselines have those type of nodes but those are not included in our generation algorithm, there is a mismatch between baseline and our data. Our goal is to extend the range of the generated data but still cover and resemble the \glspl{wdn} baselines. 
    
    \paragraph{Yearly Pattern.} The yearly pattern defines the trend of water consumption in the entire year, considering a seasonal component with a peak consumption in summer. The default configuration assumes the European summer season starting in June with a 3-month span. In addition, to introduce variability in the data, beneficial for training deep learning models, we randomly move the summer period along the entire year for approximately 20\% of simulated scenarios. This approach introduces the seasonal patterns in other regions across the globe. The yearly pattern is composed of a yearly component, a seasonal component, and noise, as described in Equation \ref{eq:yearly}. 
    
    \begin{equation}
    	\label{eq:yearly}
    	\text{yearly}(x_t) = y(x_t) + s(x_t) + z_t: ~ t \in T
    \end{equation}
    
    \noindent where $y(x_t)$ is the yearly component generated using a Fourier time-series as described by Equation \ref{eq:ycomp}, $s(x_t)$ is the seasonal pattern generated using a periodic cosine function as described by Equation \ref{eq:season}, and $z_t$ is white noise.
    
    \begin{equation}
    	\label{eq:ycomp}
    y(x_t) = A_0 + \sum_{n=1}^H \left( A_n \cos\left(2\pi \frac{n x_t}{\text{num\_samples}}\right) + B_n \sin\left(2\pi \frac{n x_t}{\text{num\_samples}}\right) \right): ~ t \in T
    \end{equation}
    
    \noindent where the Fourier coefficients $A_n$ and $B_n$ determine the amplitude of the signal, and they are randomly sampled from a uniform distribution in the range $[0,1)$, the value of $H$ represent the number of harmonics used for the time-series, and the periodicity of the signal is 24-hour for the short-term dataset and 1-year for the long-term. The periodicity is given by number of samples parameter $num\_samples$.
    
    \begin{equation}
    	\label{eq:season}
    	s(x_t) = C \left( cos\left( 2\pi \frac{x_t - s_{peak}}{\text{num\_samples}}\right) \right): ~ t \in T
    \end{equation}
    
    \noindent where C is a constant that represents the amplitude of the signal, reaching its maximum value in the summer peak $s_{peak}$, $num\_samples$ defines the periodicity of the signal. Finally, the yearly time-series are normalized in the range $[0,1]$
    
    \paragraph{Noise.} The noise component $\epsilon_t$, from Equation \ref{eq:demand_pattern}, is used to model the high and unexpected fluctuations in water consumption. Such variations can be caused by unpredictable changes in consumer behavior, network maintenance, transients or other unforeseen circumstances \cite{vrachimis2018leakdb}. The noise component was sampled from a Gaussian normal distribution centered at zero and a standard deviation randomly sampled from a uniform distribution in the range $[min\_noise, max\_noise]$.
    
    \section*{Data Records}
    The DiTEC-WDN dataset comprises 36 \glspl{wdn}, each containing 1,000 distinct scenarios. A scenario is a sequence of snapshots, capturing key measurements sampled hourly from all components. Each snapshot describes an undirected graph in which nodes involve \textit{reservoir}, \textit{tank}, and \textit{junction}, and links represent \textit{pipe}, \textit{head pump}, \textit{power pump}, \textit{PRV}, \textit{PSV}, \textit{FCV}, and \textit{TCV} valves. Note that some valve types are omitted, as they are unavailable in the dataset. In particular, we recorded input parameters of all components (as described in Table \ref{tab:parameters}) and seven simulation outputs: \textit{pressure}, \textit{demand}, \textit{head}, \textit{flow rate}, \textit{velocity}, \textit{head loss}, and \textit{friction factor} with units defined per network and in standard units. 
    
\begin{figure}[ht]
        \centering
        \includegraphics[height=0.5\textheight]{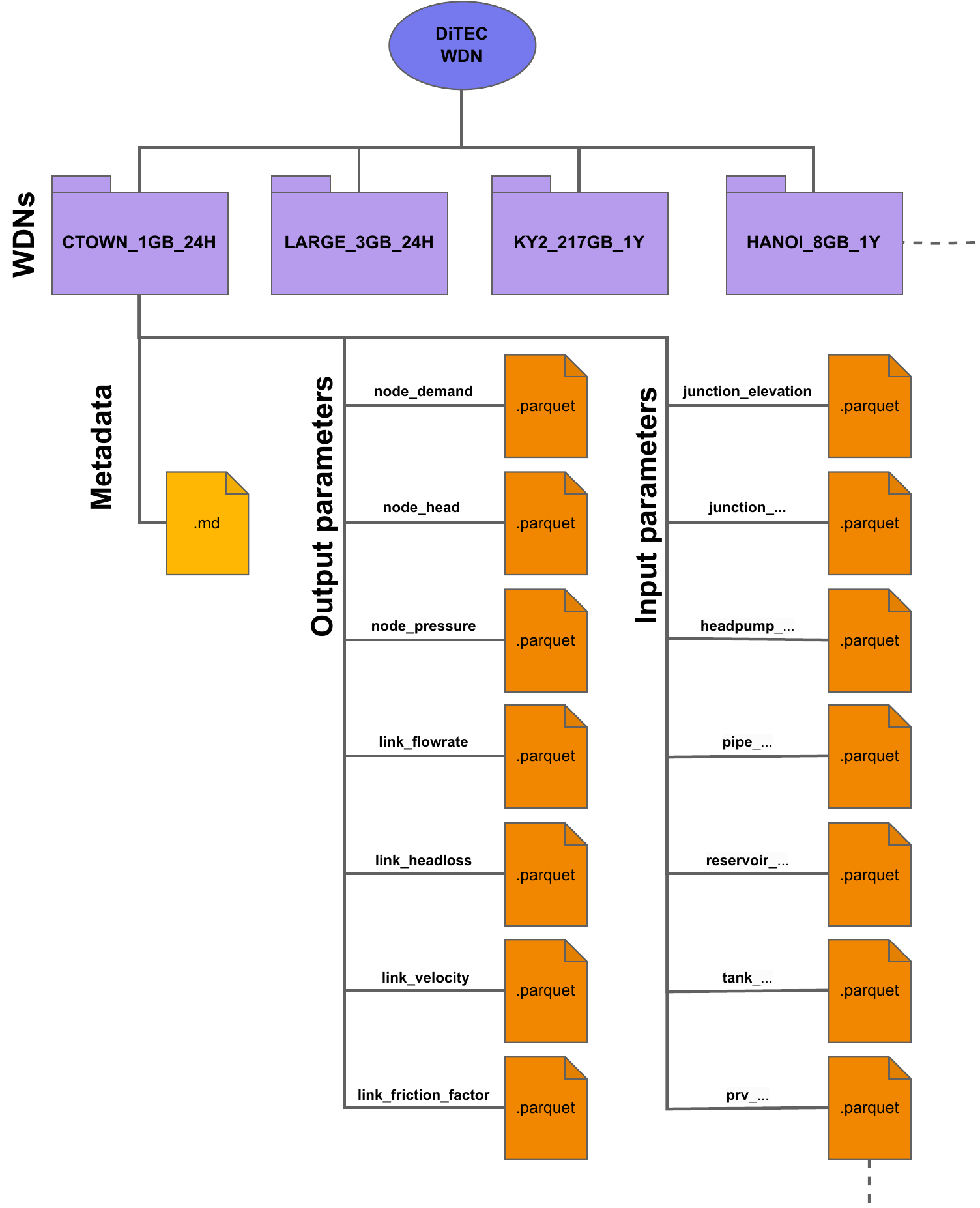}
        \caption{\textbf{The folder organization structure.} The \textbf{DiTEC-WDN} collection includes 36 \glspl{wdn} represented as folders. Every folder contains metadata and seven output parameters while the number of input parameters varies based on the available components per network. The dataset metadata is fed into a Markdown (\textit{.md}) file structured as Dataset Card~\cite{mitchell2019datasetcard}. In addition, parameter values are stored in one or more \textit{.parquet} file(s), depending on their size. A \textit{.parquet} file stores indices and node (link) values as distinct columns.}
        \label{fig:data_structure}
    \end{figure}

    Each \gls{wdn} is located in a folder named as $\textbf{<network>\_<capacity>\_<duration>}$. The $\textbf{<network>}$ name corresponds to the baseline network, the $\textbf{<capacity>}$ indicates the physical size (varying from 1 GB to 232 GB), and the $\textbf{<duration>}$ specifies the simulation period which can be 24 hours ($24H$) or 1 year ($1Y$). As shown in Figure \ref{fig:data_structure}, each directory physically contains a metadata Markdown (\textit{.md}) file, seven output parameters, and several input parameters stored in \textit{.parquet} files. The metadata includes network topology, node, edge names, and auxiliary information served for optimization, generation, and simulation phases as detailed in Table \ref{tab:zattrs}. For \textit{.parquet} files, their naming follows the syntax $\textbf{<component>\_<parameter>\_<index>\_<type>\_<io>}$. The $\textbf{<component>}$ and $\textbf{<parameter>}$ define which component the parameter belongs to. The $\textbf{<index>}$ represents the shard index of the \textit{.parquet} file. The $\textbf{<type>}$ specifies the parameter category—\textit{curve}, \textit{static}, or \textit{dynamic}—while $\textbf{<io>}$ indicates whether the parameter is \textit{input} or \textit{output}. 
    
    Each parameter is associated with a table whose values are arranged based on the parameter type as follows:
    \begin{itemize}
        \item \textit{Static} tables have dimensions $(\textit{num\_scenarios} \times \textit{num\_components})$. 
        \item \textit{Pattern} tables have dimensions $((\textit{num\_scenarios} * \textit{num\_snapshots}) \times \textit{num\_components})$. 
        \item \textit{Curve}-related tables have dimensions  
        $((\textit{num\_scenarios} * \textit{num\_curve\_points}) \times \textit{num\_components})$. 
    \end{itemize}
    where \textit{num\_scenarios} stands for the number of scenarios, \textit{num\_snapshots} represents the number of snapshots, \textit{num\_curve\_points} refers to the number of curve points, and \textit{num\_components} indicates the number of nodes or links.
    
    
    
    \section*{Technical Validation}
    \label{sec:techval}
    
    To assess the dataset quality, we compared DiTEC-WDN against (1) baseline networks and (2) LeakDB \cite{vrachimis2018leakdb}, a well-known dataset. We visualized data distribution in the former and examined \textit{demand patterns} in the latter, highlighting their scarcity and the risks of overuse in the existing dataset.
    
    \subsection*{Comparative Analysis}
    
    \paragraph{DiTEC-WDN vs. Baseline networks}
    
\begin{figure}[ht]
        \centering
        \includegraphics[height=0.5\textheight, width=0.90\textwidth]{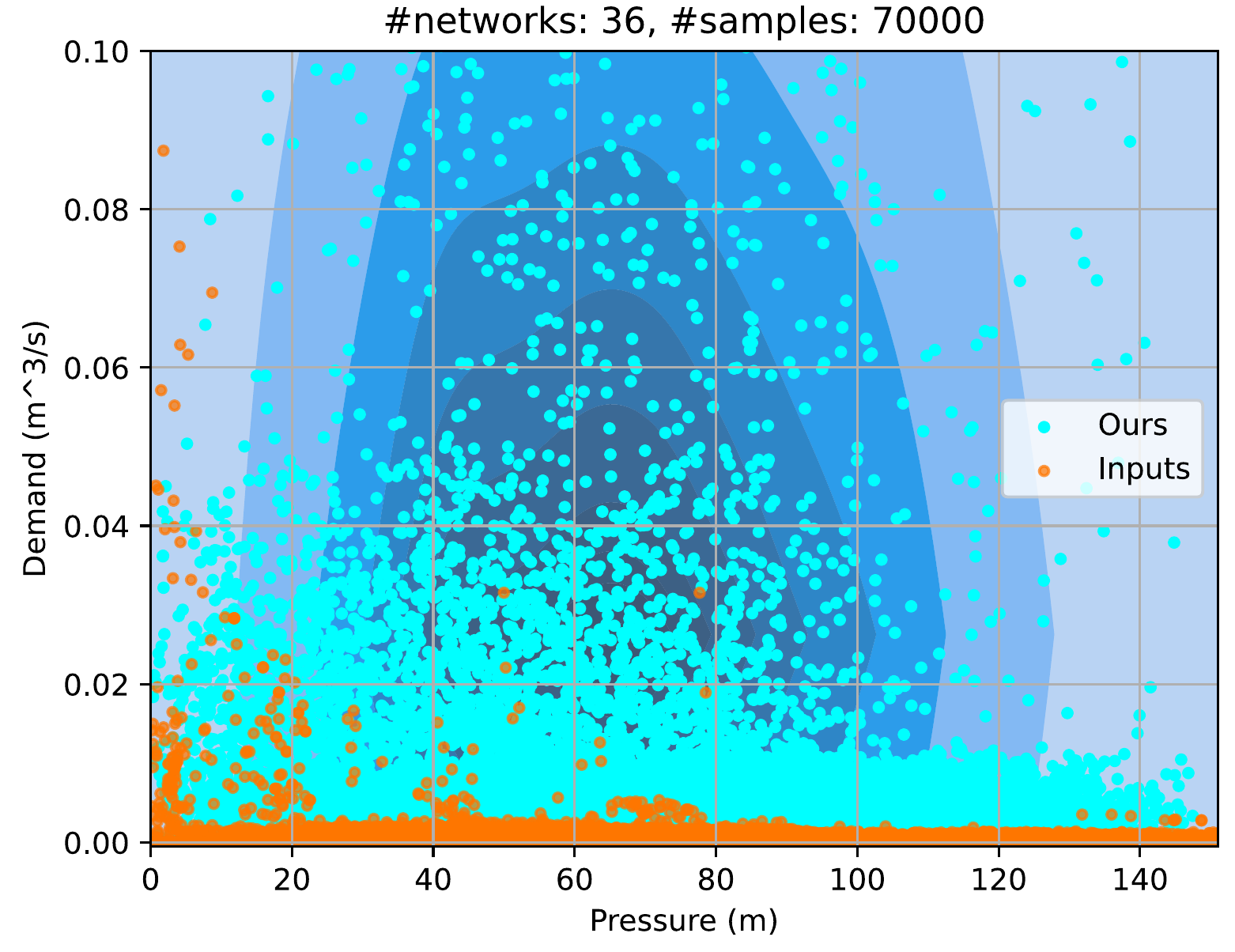}
        \caption{\textbf{Density distribution of \textit{pressure} and \textit{demand} across \glspl{wdn} in DiTEC-WDN (cyan) and original ones from Input files (orange).} The contours denote the data point density of the DiTEC-WDN dataset, with darker blue indicating higher concentration at the center and lighter blue showing lower density when going outward. In baseline networks, data points whose pressure is outside the range of $(0,151]$ in meters, are excluded due to the impractical operation conditions~\cite{truong2024gatres}.}
        \label{fig:demand_pressure_span}
    \end{figure}

    Figure \ref{fig:demand_pressure_span} highlights the contrast in data distribution between baseline networks (orange) and DiTEC-WDN dataset (cyan) along the \textit{demand} and \textit{pressure} axes. On the left, baseline data points correspond to high demand and low pressure, indicating that only a few nodes receive sufficient supply while most experience pressure drops. Similarly, on the right side, the pressure of baseline points is stable only when their corresponding demand approaches near zero. This reflects the demand scarcity and suboptimal simulation parameters. 
    An alternative approach is leveraging these networks to build a synthetic dataset, where parameters are drawn from a random distribution~\cite{tello2024gida,ashaf2024pignn, kerimov2024egnn}. However, this could violate realism and consistency. For instance, arbitrarily sampling nodal elevation or pipe diameter may result in unrealistic scenarios, such as spiky terrain or pipe bottlenecks in the \gls{wdn}. 

    In contrast, we specifically designed the parameter spaces and enforced strict rule validation to ensure hydraulic stability across scenarios while expanding into a larger space. As a result, the DiTEC-WDN dataset provides a broader, more realistic receptive field within the typically operational pressure range. This enables the robustness of training data-driven machine-learning models. Moreover, DiTEC-WDN's variability allows water researchers to analyze diverse scenarios without repetitive simulations, thereby preventing inconsistent results among studies and ensuring more sustainable research practices.

    \paragraph{DiTEC-WDN vs. LeakDB} 

\begin{figure*}
        \centering
        \begin{subfigure}[b]{0.5\textwidth}
            \centering
            \includegraphics[height=0.3\textheight]{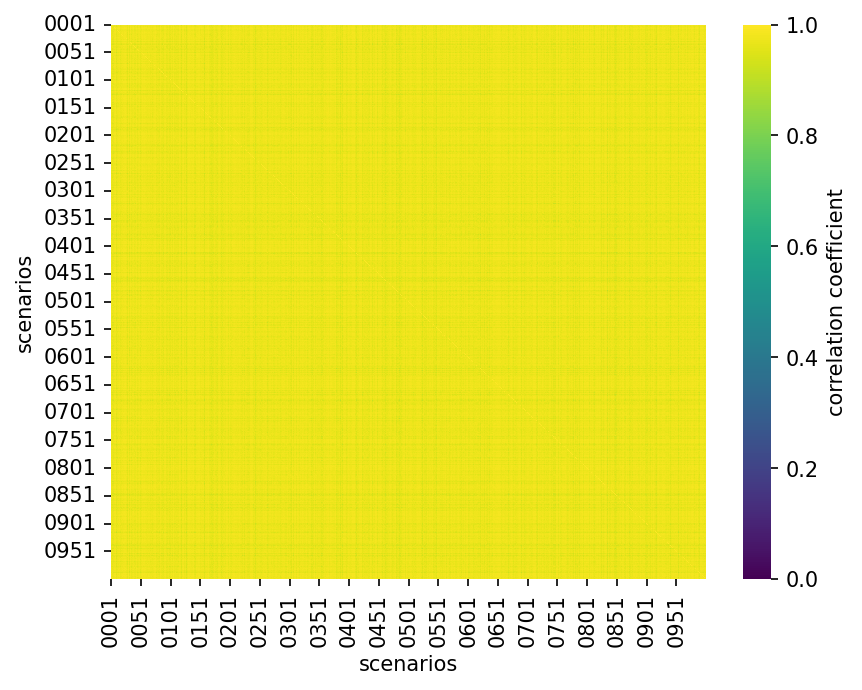}
            \caption{LeakDB}
            \label{corr-sc-leak}
        \end{subfigure}%
        ~ 
        \begin{subfigure}[b]{0.5\textwidth}
            \centering
            \includegraphics[height=0.3\textheight]{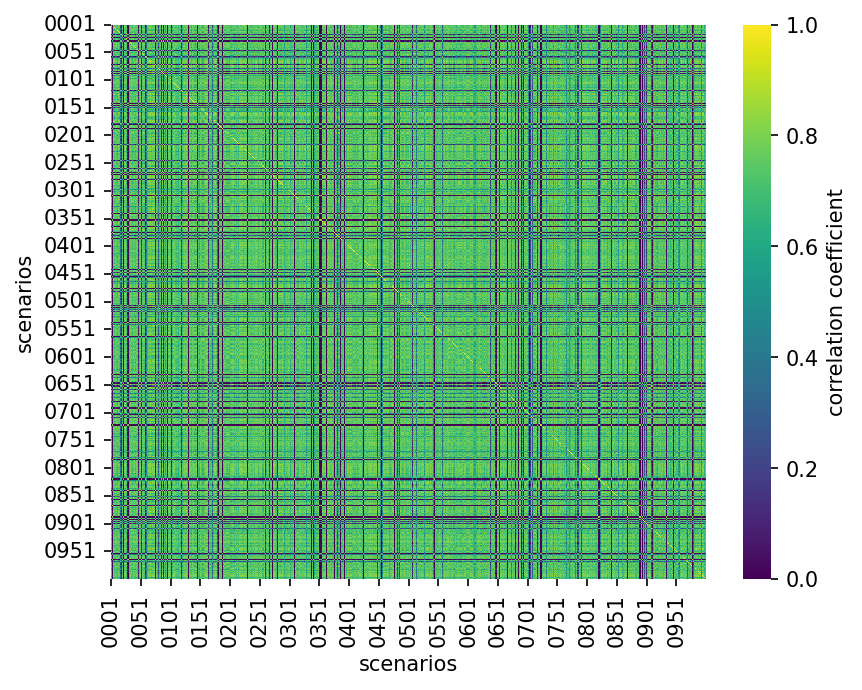}
            \caption{DiTEC-WDN}
            \label{corr-sc-ours}
        \end{subfigure}
        \caption{\textbf{Correlation matrices of generated demands between all scenarios in Hanoi \gls{wdn}}. The left figure (a) shows the correlation between scenarios in the data generated in LeakDB \cite{vrachimis2018leakdb}. The right figure (b) shows the correlation between the scenarios in our dataset. Both matrices include all 1,000 scenarios, each containing 1-year of demand data. The low correlation between scenarios in our dataset shows the diversity of the data, contrary to the similarity observed across LeakDB scenarios.}
        \label{fig:corr-scenarios}
    \end{figure*}

    Another important analysis is how our generated data differ from the commonly-used existing benchmark dataset, LeakDB\cite{vrachimis2018leakdb}. Figure \ref{fig:corr-scenarios} shows the demand correlation matrices between the 1,000 scenarios generated in LeakDB and our generated data. As can be seen in Figure \ref{corr-sc-leak} the scenarios generated in LeakDB are highly correlated. The correlation matrix shows only slight variations between some scenarios, which implies data redundancy. This limits the capacity of deep learning models to learn from such data. On the contrary, in our dataset, the correlation between scenarios is much lower as can be seen in Figure \ref{corr-sc-ours}. This confirms the diversity of the generated data, allowing the models to see a larger space of solutions during the training process.    

\begin{figure*}
        \centering
        \begin{subfigure}[b]{0.5\textwidth}
            \centering
            \includegraphics[height=0.3\textheight]{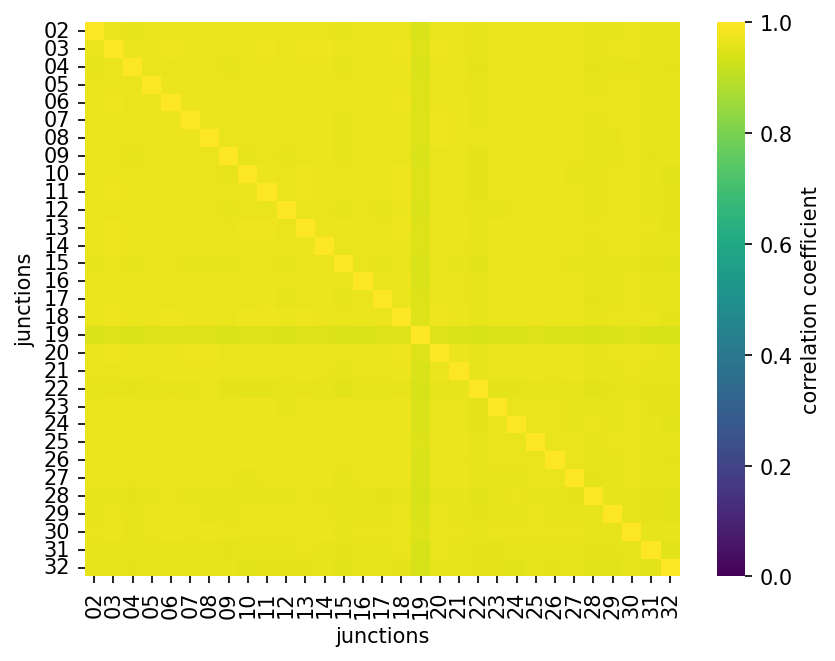}
            \caption{LeakDB}
            \label{fig:corr-n-leak}
        \end{subfigure}%
        ~ 
        \begin{subfigure}[b]{0.5\textwidth}
            \centering
            \includegraphics[height=0.3\textheight]{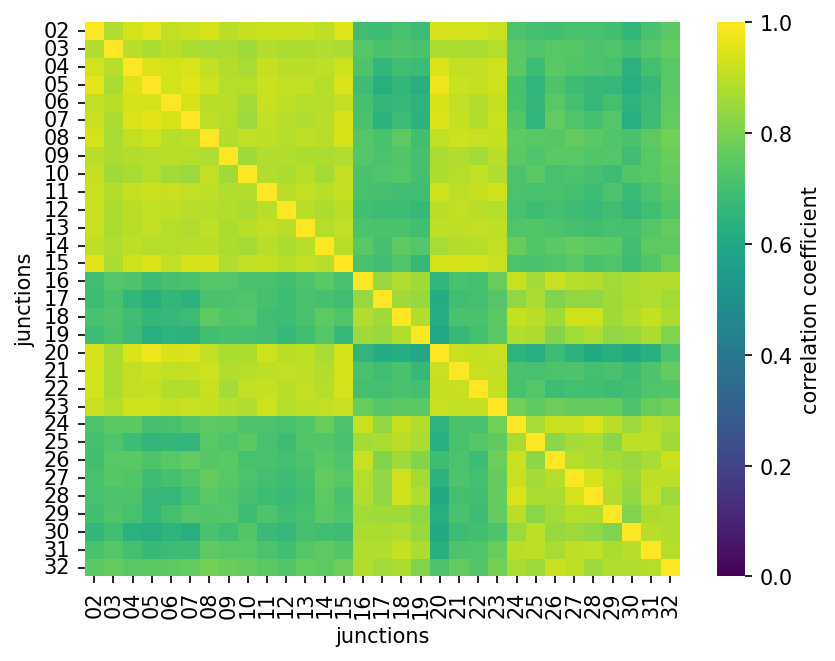}
            \caption{DiTEC-WDN}
            \label{fig:corr-n-ours}
        \end{subfigure}
        \caption{\textbf{Correlation matrices of generated demands between junction nodes in a randomly chosen scenario from Hanoi \gls{wdn}}. The left figure (a) shows the correlation between junction demands in the data generated in LeakDB \cite{vrachimis2018leakdb}. The right figure (b) shows the correlation between the junction demands in our dataset. The high correlation in LeakDB shows the overuse of demand patterns for several nodes, contrary to what it is observed in our dataset. The blocks in the correlation matrix of our dataset highlight the difference between household and commercial demand patterns.}
        \label{fig:corr-nodes}
    \end{figure*}

    Similar conclusions can be drawn from the correlation matrices between the junction demands in an arbitrary scenario (see Figure \ref{fig:corr-nodes}). The high correlation shown in Figure \ref{fig:corr-n-leak} exposes the negative effect of demand patterns overuse in existing approaches. In contrast, Figure \ref{fig:corr-n-ours} shows a moderate correlation between junction demands in our data, implying there is a pattern in consumption demand, but this is not identical for every node in the \gls{wdn}. In addition, the block patterns shown in Figure \ref{fig:corr-n-ours} display the difference between households and commercial consumption profiles described in Section Automatic Demand Generator. 

\begin{figure}
        \centering
        \begin{subfigure}[b]{0.95\textwidth}
            \centering
            \includegraphics[width=\textwidth]{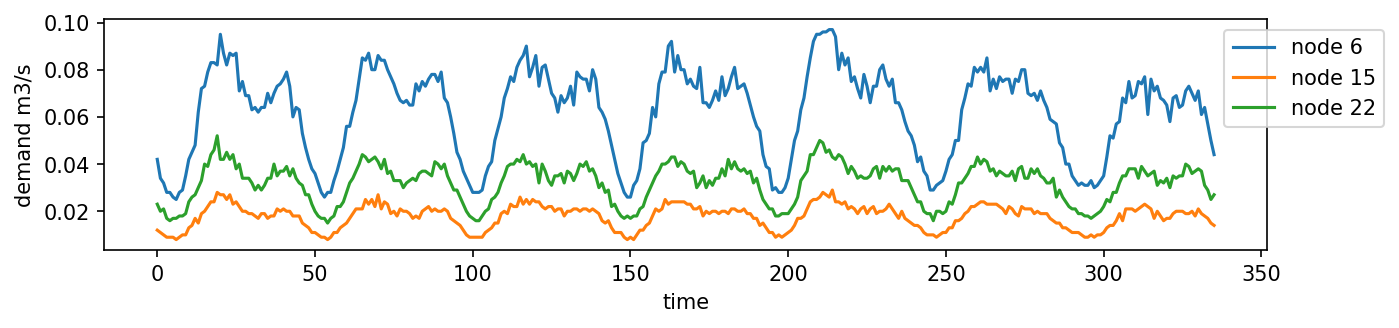}
            \caption{LeakDB}
            \label{fig:ts-leak}
        \end{subfigure}
        \begin{subfigure}[b]{0.95\textwidth}
            \centering
            \includegraphics[width=\textwidth]{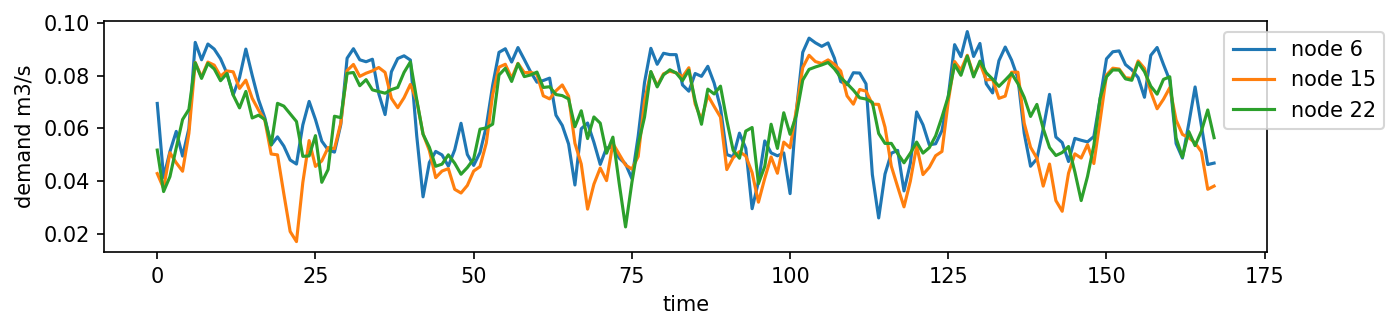}
            \caption{DiTEC-WDN}
            \label{fig:ts-ours}
        \end{subfigure}
        \caption{\textbf{Time series of the generated demands of three randomly chosen junction nodes from Hanoi \gls{wdn}}. Figure (a) shows one week of the demands generated in LeakDB \cite{vrachimis2018leakdb}, sampled every 30 minutes. The reuse of a single pattern for different nodes is clearly observed in LeakDB. Figure (b) shows one week of demands from our dataset, sampled every 60 minutes. In this case, the fluctuations observed in the time series show a different consumption pattern per node.}
        \label{fig:demands-ts}
    \end{figure}

    Finally, the time series of one week demand for three nodes from a random scenario in LeakDB and our dataset are shown in Figure \ref{fig:demands-ts}. The time series depicted in Figure \ref{fig:ts-leak} show how a single demand pattern is reused for the three nodes in LeakDB. While the noise shows some subtle variations, each time series looks like a translated and scaled version of the other. Contrary, our data exhibit consumption patterns, but the fluctuations in each time series are independent, as shown in Figure \ref{fig:ts-ours}.

    \section*{Usage Notes}
   
    \paragraph{Privacy and safety control}
    The DiTEC-WDN dataset is available at \href{https://huggingface.co/datasets/rugds/ditec-wdn}{https://huggingface.co/datasets/rugds/ditec-wdn} under CC BY-NC 4.0 license. This dataset comprises 36,000 synthesized snapshots devised from publicly available \glspl{wdn} that served as structural backbones. Specifically, the network's topology, node names, and link names remain unchanged, while other parameter values are machine-generated automatically.

    \paragraph{Data loading}
    The repository where the raw dataset is located supports several data interface options to read and process \textit{.parquet} files, allowing practitioners to select a concrete parameter or a subset of networks. Before use, the downloaded dataset requires an additional preprocessing step. Specifically, we removed columns corresponding to nodes along with their adjacent links, listed in \textit{skip\_names} in the metadata. Network metadata is accessible in any \textit{.parquet} file in the corresponding folder. Additionally, to analyze graph topology, the metadata contains \textit{adj\_list} formatted as a list of tuples (source node, adjacent link, destination node).
    
    \paragraph{Limitations}
    Despite the diversity of simulation parameters recorded in the dataset, it is important to note some limitations in replicating unexpected situations and storing the water flow direction and auto-generated hyper-parameters. We assume all scenarios are under normal conditions and components are functioning correctly. Accordingly, anomaly occurrences, such as negative pressure, leakage, fire-fighting, or pipe break, are excluded from the dataset. 
    
    For the second limitation, the integration of flow direction demands a one-by-one mapping between the network topology and each scenario and, therefore, increases the storage requirement exponentially. Conversely, we used a shared topology as an undirected graph for all scenarios within the same \gls{wdn} to save computational units. Still, the consequence trade-off is the incompatibility of the dataset with direction-related tasks such as flow pattern prediction, fracture flow, and fluid analysis. To compensate for that, a possible solution is to transform the input simulation parameters to a customized \textit{.INP} file and leverage the simulation tool such as EPANET~\cite{rossman1999epanet} to re-generate the flow directions.
    
    In line with this, some hyper-parameters generated during the simulation process, such as the locations of extreme-demand and zero-demand nodes, and nodal demand profiles, cannot be recorded. To address this, these extreme-demand and zero-demand nodes can be identified using high-pass and low-pass filters, respectively, while demand profilers can be classified by an unsupervised machine-learning algorithm, such as K-Means~\cite{lloyd1982kmean} or DBSCAN~\cite{ester1996dbscan}.
    
    \section*{Code availability}
    \label{sec:code_avail}
    The optimization algorithm and generation tool are written in Python and available on \href{https://github.com/DiTEC-project/DiTEC_WDN_dataset}{Github} under MIT License. The repository includes a detailed tutorial and wiki to guide scenario generation for a customized \gls{wdn}. The outcome dataset is stored in \textit{Zarr}, an efficient compressed format. Conversion to \textit{.parquet} can be performed using the \textit{zarr2parquet.py} script.
    
    \bibliography{bibliography}
    
    
    \section*{Acknowledgements} 
    
    This work is funded by the project DiTEC: Digital Twin for Evolutionary Changes in Water Networks (NWO 19454). We thank the Center for Information Technology of the University of Groningen for their support and for providing access to the Hábrók high performance computing cluster. Also, we appreciate Hugging Face for hosting the dataset repository.
    
    \section*{Author contributions statement}

    All authors conceptualized the idea. H.T., A.T. and V.D. involved in writing and reviewing this manuscript. H.T. contributed to the methodology, developed the optimization and generation tool, investigation, visualization, and data curation. A.T. contributed in methodology, validation, visualization, and developed the demand generation. A.L. and V.D. provided resources, supervised the project, and contributed to project administration and funding acquisition.

    \section*{Competing interests} 
    
    The authors declare no competing interests.
    
    \section*{Figures \& Tables}
    
    
    
    
    
    
    \clearpage

    \clearpage

            \renewcommand{\arraystretch}{1.2}
    	\begin{longtable}{|m{2.4cm}|m{5.1cm}|c|c|c|c|c|c|}
    		\hline \hline
    		\textbf{WDN} & \textbf{Description} & \textbf{Junctions} & \textbf{Pipes} & \textbf{Reservoirs} & \textbf{Tanks} & \textbf{Pumps} & \textbf{Patterns} \\ \hline
             \endfirsthead

            \multicolumn{4}{p{10cm}}{\scriptsize {\tablename} \thetable{} -- Continued} \\[0.5ex]
              \hline \\[-2ex]
    		\textbf{WDN} & \textbf{Description} & \textbf{Junctions} & \textbf{Pipes} & \textbf{Reservoirs} & \textbf{Tanks} & \textbf{Pumps} & \textbf{Patterns} \\ \hline
            \endhead

            \multicolumn{4}{p{10cm}}{\scriptsize{Continues on Next Page\ldots}} \\
            \endfoot
            
            \endlastfoot
             
    		ky1 \cite{jolly2014research} & \multirow{14}{*}{\parbox{4.5cm}{Synthetic \glspl{wdn} based on a statewide database of water systems originally developed by the Kentucky Infrastructure Authority, in United States.}} & 856 & 985 & 1 & 2 & 1 & 2 \\  \cline{1-1} \cline{3-8}
    		 ky2 \cite{jolly2014research} & & 811 & 1125 & 1 & 3 & 1 & 3 \\ \cline{1-1} \cline{3-8}
    		 ky3 \cite{jolly2014research} & & 269 & 371 & 3 & 3 & 5 & 3 \\ \cline{1-1} \cline{3-8}
    		 ky4 \cite{jolly2014research} & & 959 & 1158 & 1 & 4 & 2 & 3 \\ \cline{1-1} \cline{3-8}
    		 ky5 \cite{jolly2014research} & & 420 & 505 & 4 & 3 & 9 & 3 \\ \cline{1-1} \cline{3-8}
    		 ky6 \cite{jolly2014research} & & 543 & 647 & 2 & 3 & 2 & 4 \\ \cline{1-1} \cline{3-8}
    		 ky7 \cite{jolly2014research} & & 481 & 604 & 1 & 3 & 1 & 4 \\ \cline{1-1} \cline{3-8}
    		 ky8 \cite{jolly2014research} & & 1325 & 1618 & 2 & 5 & 4 & 4 \\ \cline{1-1} \cline{3-8}
    		 ky10 \cite{jolly2014research} & & 920 & 1061 & 2 & 13 & 13 & 4 \\ \cline{1-1} \cline{3-8}
    		 ky13 \cite{jolly2014research} & & 778 & 944 & 2 & 5 & 4 & 3 \\ \cline{1-1} \cline{3-8}
    		 ky14 \cite{jolly2014research} & & 377 & 553 & 4 & 3 & 5 & 3 \\ \cline{1-1} \cline{3-8}
    		 ky16 \cite{jolly2014research} & & 791 & 915 & 3 & 4 & 7 & 3 \\ \cline{1-1} \cline{3-8}
    		 ky18 \cite{jolly2014research} & & 772 & 917 & 4 & 0 & 3 & 9 \\ \cline{1-1} \cline{3-8}
    		 ky24\_v \cite{jolly2014research} & & 288 & 292 & 2 & 0 & 0 & 3 \\ \hline             
    		19 Pipe System \cite{wood1972hydraulic} & An artificial WDN with two sources. It was used for analyzing flow distribution in hydraulic networks. & 12 & 21 & 2 & 0 & 0 & 3 \\ \hline
    		Anytown \cite{walski1987battle} & It is a hypothetical WDN used as part of a Battle of the Networks competition aimed at improving analysis methods.  & 19 & 41 & 3 & 0 & 1 & 1 \\ \hline
    		new\_york \cite{schaake1969linear} & It represents the water supply transmission tunnels for the City of New York in 1969. It was originally used to optimize duplications to the existing system to meet projected demand increases. & 19 & 42 & 1 & 0 & 0 & 4 \\ \hline
    		Jilin \cite{bi2014optimization} & It is a synthetic network used as part of a study of optimization of WDNs via online retrained metamodels. & 27 & 34 & 1 & 0 & 0 & 1 \\ \hline
    		hanoi \cite{fujiwara1990two} & This WDN is based on the planned trunk network of Hanoi, Vietnam. It was originally used to test pipe size optimization software. & 31 & 34 & 1 & 0 & 0 & 0 \\ \hline
    		fossolo \cite{bragalli2008ibm} (foss\_poly\_1) & It is based on the WDN of the Fossolo neighborhood in Bologna, Italy. It was used for WDN design optimization. & 36 & 58 & 1 & 0 & 0 & 0 \\ \hline
    		FOWM \cite{walski05} & It is a skeletonized version of the WDN of northern Arlington County in the United States. & 44 & 49 & 1 & 0 & 0 & 0 \\ \hline
    		EPANET Net 3 \cite{clark1995modeling} & It is based on the North Marin WDN in Novato, California. It was used as part of a water quality study.  & 92 & 119 & 2 & 3 & 2 & 5 \\ \hline
    		FFCL-1 \cite{rossman1996numerical} & It is based on the Fairfield WDN, a relatively small system with a single source. It was originally used to study numerical modeling methods for water quality.  & 111 & 126 & 0 & 1 & 0 & 3 \\ \hline
    		Zhi Jiang \cite{zheng2011combined} (ZJ) & It is a simplified version of the Zhi Jiang WDN in the eastern province of China. It was originally used as part of a design and optimization study. & 113 & 164 & 1 & 0 & 0 & 0 \\ \hline
    		WA1 \cite{vasconcelos1997kinetics} & It is based on the Bellingham WDN in Washington, US. It was originally used for water quality modeling. & 121 & 169 & 0 & 2 & 0 & 6 \\ \hline
    		OBCL-1 \cite{vasconcelos1997kinetics} & It is based on the Cheshire WDN near located Harrisburg, Pennsylvania. It was used originally to study the kinetics of chlorine decay. & 262 & 289 & 1 & 0 & 1 & 5 \\ \hline
    		modena \cite{bragalli2008ibm} & It is a simplified version of the WDN of the town of Modena, Italy. It was originally used for WDN design studies. & 268 & 317 & 4 & 0 & 0 & 0 \\ \hline
    		NPCL-1 \cite{clark1994applying} & It is based on the North Penn Water Authority WDN. It was original used for water quality studies. & 337 & 399 & 0 & 2 & 0 & 17 \\ \hline
    		Marchi Rural \cite{marchi2014methodology} (RuralNetwork) & This WDN was adapted from an irrigation system in Australia. It was originally used as part of a design and optimization study. & 379 & 476 & 2 & 0 & 0 & 0 \\ \hline
    		CTOWN \cite{ostfeld2012battle} & It is based on a real small \gls{wdn}, the data was in part obtained from a  geographic information system of the Municipality of C-Town, and part from SCADA systems. & 388 & 444 & 1 & 7 & 11 & 5 \\ \hline
    		d-town \cite{marchi2014battle} & It is a hypothetical WDN created as part of a Battle of the Networks focused on long term improvement plans that account for greenhouse gas emissions. & 399 & 459 & 1 & 7 & 11 & 5 \\ \hline
    		balerma \cite{reca2006genetic} & This WDN is an adaption of an existing irrigation network in the Sol-Poniente irrigation district, located in Balerma in the province of Almer\'{i}a in Spain.  & 443 & 454 & 4 & 0 & 0 & 0 \\ \hline
    		L-TOWN \cite{vrachimis2020dataset} & It is a synthetic WDN, based on a real WDN of a city in Cyprus. It was created for The Battle of Leakage Detection and Isolation Methods (BattLeDIM). & 782 & 909 & 2 & 1 & 1 & 107 \\ \hline
    		KL \cite{kang2012revisiting} & It is a synthetic WDN, originally used in a study on the heuristic hierarchical approach to optimization of WDN design. & 935 & 1274 & 1 & 0 & 0 & 0 \\ \hline
    		Exnet \cite{farmani2004exnet} (EXN) & It is a synthetic WDN proposed by the Centre for Water Systems of Exeter University. It was created as a benchmark in multi-objective optimization problems. & 1891 & 2467 & 2 & 0 & 0 & 0 \\ \hline
    		Large \cite{sitzenfrei2023dual} & It is a hypothetical WDN based on a real data. It has one source node, which supplies the entire WDN. It was used originally for design optimization. & 3557 & 4021 & 1 & 0 & 0 & 0 \\ \hline \hline
    	\caption{List of collected \glspl{wdn}.}
    	\label{tab:baselines}
        \end{longtable}

    \begin{table}[ht]
    \centering
    \begin{tabular}{|>{\centering\arraybackslash}p{5cm}|c|c|>{\centering\arraybackslash}p{1.8cm}|l|}
    \hline
     \textbf{Components} & \textbf{Parameter} & \textbf{Type} & \textbf{Unit} & \textbf{Global Range/States}\\
    \hline
    Head pump, Power pump, Pipe, PRV, PSV, FCV, TCV & Initial Status & Static (Category)  & -  & Closed/Opened/Active/CV \\\hline
    Head pump, Power pump & Base speed & Static (Float) & -  & [0.9, 1.0] \\\hline
    Head pump, Power pump & Efficiency X & Curve &  SIFU$^{a}$  & [0.0, 0.5] \\\hline
    Head pump, Power pump & Efficiency Y & Curve & \%  & [0.0, 77.0] \\\hline
    Head pump & Pump curve X & Curve & SIFU & [0.0, 0.88] \\\hline
    Head pump & Pump curve Y & Curve & m & [0.0, 211.02] \\\hline
    Head pump & Energy pattern & Pattern & kW-hours & [0.024093, 0.1234] \\\hline
    Power pump & Power & Static (Float) & kW & [372.85, 186424.97] \\\hline
    Pipe & Diameter & Static (Float) & mm &  [0.0010, 5.1816]     \\\hline
    Pipe & Minor loss & Static (Float) & -  & [0,1000]\\\hline
    Pipe & Roughness & Static (Float) & mm (DW$^{b}$) \newline - (Otherwise) & [0.0015, 8333.3333]\\\hline
    Pipe & Length & Static (Float) & m & [0.01, 17003.20] \\\hline
    PRV & Initial Setting & Static (Float) & m & [0.0, 154.75] \\\hline
    PSV & Initial Setting & Static (Float) & m & [38.69, 49.23] \\\hline
    FCV & Initial Setting & Static (Float) & SIFU & [0.0, 0.9] \\\hline
    TCV & Initial Setting & Static (Float) & - & [0.0, 403101800000] \\\hline
    Tank & Elevation & Static (Float) & m & [2.00, 571.12]\\\hline
    Tank & Diameter & Static (Float) & m & [0.3048, 58.309]\\\hline
    Tank & Initial level & Static (Float) & m &  [0.50, 548.64]\\\hline
    Tank & Minimum volume & Static (Float) & m$^3$ &  [0.000, 95965.597]\\\hline
    Junction & Input demand & Pattern & SIFU & [-1.388, 4.814]\\\hline
    Junction & Elevation & Static (Float) & m & [0., 154.75]\\\hline
    Reservoir & Base head & Static (Float) & m & [0, 500]\\\hline
    Reservoir & Head pattern & Pattern & m  & [0.91, 70.42]\\\hline
    
    \multicolumn{5}{l}{$^{a}$ SIFU stands for SI Flow Units including LPS, LPM, MLD, CMH, and CMD. }\\
    \multicolumn{5}{l}{$^{b}$ DM refers to Darcy Weisbach headloss equation.}
    \end{tabular}
    \caption{\label{tab:parameters} List of available hydraulic parameters.}
    
    \end{table}
    
    \begin{table}[ht]
    \centering
    \footnotesize
    \begin{tabular}{|l|c|>{\arraybackslash}p{10cm}|}
    \hline
     \textbf{Metadata} & \textbf{Data Type} & \textbf{Description}  \\
    \hline
    adj\_list & List & Adjacency list in which each element has a format of (source node name, dest node name, pipe name). \\\hline
    backup\_times & Float & Backup time.\\\hline
    batch\_size & Integer & Batch size defines how many samples the simulation takes per time. \\\hline
    duration & Integer & Simulation time in hours. \\\hline
    extreme\_dem\_rate & Float & Extreme demand rate indicates the rate appearing extreme demand in some nodes. \\\hline
    fcv\_tune & Dict & FCV's optimized configuration. \\\hline
    fractional\_cpu\_usage & Float & Settings the CPU usage per worker as a part of the optimization process. \\\hline
    gen\_batch\_size & Integer & Batch size of random matrix generation, a part of the simulation process. \\\hline
    gpv\_tune & Dict & GPV's optimized configuration.  \\\hline
    head\_pump\_tune & Dict & Head pump's optimized configuration.  \\\hline
    index\_tracers & List & Selected scenario indices for recovering an interrupted simulation.  \\\hline
    inp\_paths & List & Paths to .INP file containing metadata of a \gls{wdn}.  \\\hline
    junction\_tune & Dict & Junction's optimized configuration.  \\\hline
    max\_extreme\_dem\_junctions & Integer & The maximum allowed amount of extreme nodes in a scenario.\\\hline
    mem\_per\_worker & Float & Allocated memory in GB for each worker in simulation process.\\\hline
    mem\_per\_worker & Float & Allocated memory in GB for each worker in simulation process.\\\hline
    noise\_range & Tuple & Lower and upper bounds of the addition noise in generating demand patterns.\\\hline
    num\_cpus & Integer & Number of CPUs dedicated for the simulation process.\\\hline
    num\_samples & Integer & Number of expected scenarios.\\\hline
    odims & Ordered Dict & parameter dimension associated with available components\\\hline
    okeys & Ordered Dict & parameter names associated with available components\\\hline
    onames & Ordered Dict & instance names associated with available components\\\hline
    output\_path & String & path storing the simulation outcome.\\\hline
    p\_commercial & Tuple & Lower and upper bounds of the demand of commercial nodes in generating demand patterns. \\\hline
    pbv\_tune & Dict & PBV's optimized configuration.   \\\hline
    pipe\_tune & Dict & Pipe's optimized configuration.   \\\hline
    power\_pump\_tune & Dict & Power pump's optimized configuration.   \\\hline
    pressure\_range & Tuple & Lower and upper bounds of a valid pressure.   \\\hline
    profile\_commercial& Tuple & Demand level of four quarters of the day in a commercial node.  \\\hline
    profile\_extreme& Tuple & Demand level of four quarters of the day in an extreme node.  \\\hline
    profile\_household& Tuple & Demand level of four quarters of the day in a household node.  \\\hline
    prv\_tune & Dict & PRV's optimized configuration.   \\\hline
    psv\_tune & Dict & PSV's optimized configuration.   \\\hline
    ray\_temp\_path & String & Temporarily path for Ray.   \\\hline
    reservoir\_tune & Dict & Reservoir's optimized configuration.  \\\hline
    save\_success\_inp & Boolean & Flag indicates whether saving a valid scenario in INP file for debugging only. \\\hline
    sim\_outputs & List & Simulation outputs to be recorded \\\hline
    skip\_names & List & Some abnormal nodes should be skipped in validation stage.\\\hline
    summer\_amplitude\_range & Tuple & Amplitude range of demand increase during summer period\\\hline
    summer\_rolling\_rate & Float & Probability of rolling summer period to mimic opposite seasons between two hemispheres.\\\hline
    summer\_start & Float & normalized time remarking the beginning of summer.\\\hline
    tank\_tune & Dict & Tank's optimized configuration. \\\hline
    tcv\_tune & Dict & TCV's optimized configuration. \\\hline
    temp\_path & String & Path to a folder storing temporary files. \\\hline
    time\_consistency & Boolean & Flag indicates whether the input and output time series must be equal in length.\\\hline
    time\_step & Float & Simulated time sampling rate in hours.\\\hline
    verbose & Boolean & Flag indicates whether to print debug information during the simulation process.\\\hline
    yearly\_pattern\_num\_harmonics & Integer & The number of terms in a Fourier series for a yearly pattern.\\\hline
    yield\_worker\_generator & Boolean & Flag indicates a generator to yield simulation outputs for saving memory.\\\hline
    zero\_dem\_rate & Float & Probability of appearing zero-demand nodes serving as water flow transitions and connections in the water system.\\\hline
    \end{tabular}
    \caption{\label{tab:zattrs} List of metadata recorded in the \textbf{.md} file.}
    
    \end{table}
    
    
    \end{document}